\documentclass[acmsmall]{acmart}

\usepackage{hyperref}
\usepackage{footnotebackref}
\usepackage{enumitem}
\usepackage{graphicx}
\usepackage{algorithm}
\usepackage{bm}
\usepackage{amsmath}
\usepackage{amsthm}
\usepackage{amsfonts}
\usepackage{longtable}
\usepackage{booktabs}
\usepackage{url}
\usepackage{capt-of}
\usepackage{xcolor}
\usepackage[noend]{algpseudocode}
\usepackage{subfig}

\algblock{Input}{EndInput}
\algnotext{EndInput}
\algblock{Output}{EndOutput}
\algnotext{EndOutput}

\newcommand{\Desc}[2]{\State \makebox[2em][l]{#1}#2}

\theoremstyle{definition}
\newtheorem{definition}{Definition}[section]

\AtBeginDocument{%
  }

 \acmJournal{TOIT}

\begin{document}

\title{Positional Encoding-based Resident Identification in Multi-resident Smart Homes}

\author{Zhiyi Song}
\email{zson5784@uni.sydney.edu.au}
\affiliation{%
  \institution{The University of Sydney}
  \city{Sydney}
  \country{Australia}
}

\author{Dipankar Chaki}
\email{dipankar.chaki@sydney.edu.au}
\affiliation{%
  \institution{The University of Sydney}
  \city{Sydney}
  \country{Australia}
}

\author{Abdallah Lakhdari}
\email{abdallah.lakhdari@sydney.edu.au}
\affiliation{%
  \institution{The University of Sydney}
  \city{Sydney}
  \country{Australia}
}

\author{Athman Bouguettaya}
\email{athman.bouguettaya@sydney.edu.au}
\affiliation{%
  \institution{The University of Sydney}
  \city{Sydney}
  \country{Australia}
}

\begin{abstract}
    We propose a novel resident identification framework to identify residents in a multi-occupant smart environment. The proposed framework employs a feature extraction model based on the concepts of positional encoding. The feature extraction model considers the locations of homes as a graph. We design a novel algorithm to build such graphs from layout maps of smart environments. The Node2Vec algorithm is used to transform the graph into high-dimensional node embeddings. A Long Short-Term Memory (LSTM) model is introduced to predict the identities of residents using temporal sequences of sensor events with the node embeddings. Extensive experiments show that our proposed scheme effectively identifies residents in a multi-occupant environment. Evaluation results on two real-world datasets demonstrate that our proposed approach achieves 94.5\% and 87.9\% accuracy, respectively.
\end{abstract}





\maketitle

\section{Introduction}

Internet of Things (IoT) refers to physical objects equipped with sensors, software, and computing power that communicate with other systems and devices over the Internet \cite{atzori2010internet}. IoT is emerging due to the rapid advancement of underlying technologies such as Radio Frequency Identification (RFID), Near Field Communication (NFC), wired sensor networks, and wireless sensor networks \cite{10.1145/3464960}. IoT technologies have been the driving force behind prominent applications such as \emph{smart campuses}, \emph{smart offices}, \emph{smart cities}, \emph{intelligent transport systems}, and \emph{smart grids} \cite{6099519}. \emph{Smart home} is another cutting-edge application of IoT. A smart home is any regular home fitted with various IoT devices \cite{10.1145/3301551.3301575}. These IoT devices are attached to everyday ``things" to monitor usage patterns. For example, a sensor (i.e., an IoT device) attached to a cup may monitor a resident's tea cup usage patterns. The IoT paradigm brings enormous opportunities to smart homes to make residents' home life more \emph{convenient} and \emph{efficient} \cite{10.1145/3301551.3301575,huang2018convenience}.

The advent of intelligent technologies such as artificial intelligence, predictive analytics, and machine learning may enable \emph{smart services} in the home environment by automating IoT devices \cite{amiribesheli2015review}. Such smart home services may help elderly and disabled people live with less reliance on others with activities of daily living \cite{do2018rish, elderly2019fall}. Residents can remotely control these devices and customize IoT-based applications via various tools and platforms such as Samsung SmartThings \cite{Samsung}. In addition, IoT-based applications can be developed employing the trigger-action paradigm \footnote{https://ifttt.com/home} (e.g., IFTTT). An example of such a rule is, ``If the TV is turned on, turn off the light". However, the current services require residents to manually set these rules. Furthermore, residents' preferences typically change and vary over time \cite{UsersTemporalPreference}. Consequently, adjustments are required to tune these rules. These cumbersome tasks, such as frequent adjustments, may cause residents to lose confidence in smart home systems and stop using them \cite{CoPI}. In this regard, a smart home system may have the ability to learn inhabitants' preferences without their intervention and adjust appliances' settings accordingly \cite{wang2010adaptive}. This may reduce the human labor and effort in producing services, thus, \emph{improving residents' quality of home life}.

\begin{figure}[!h]
    \centering
    \subfloat[Example: layout map of a simple room with a temporal sequence of sensor events.]{
        \includegraphics[width=6.8cm]{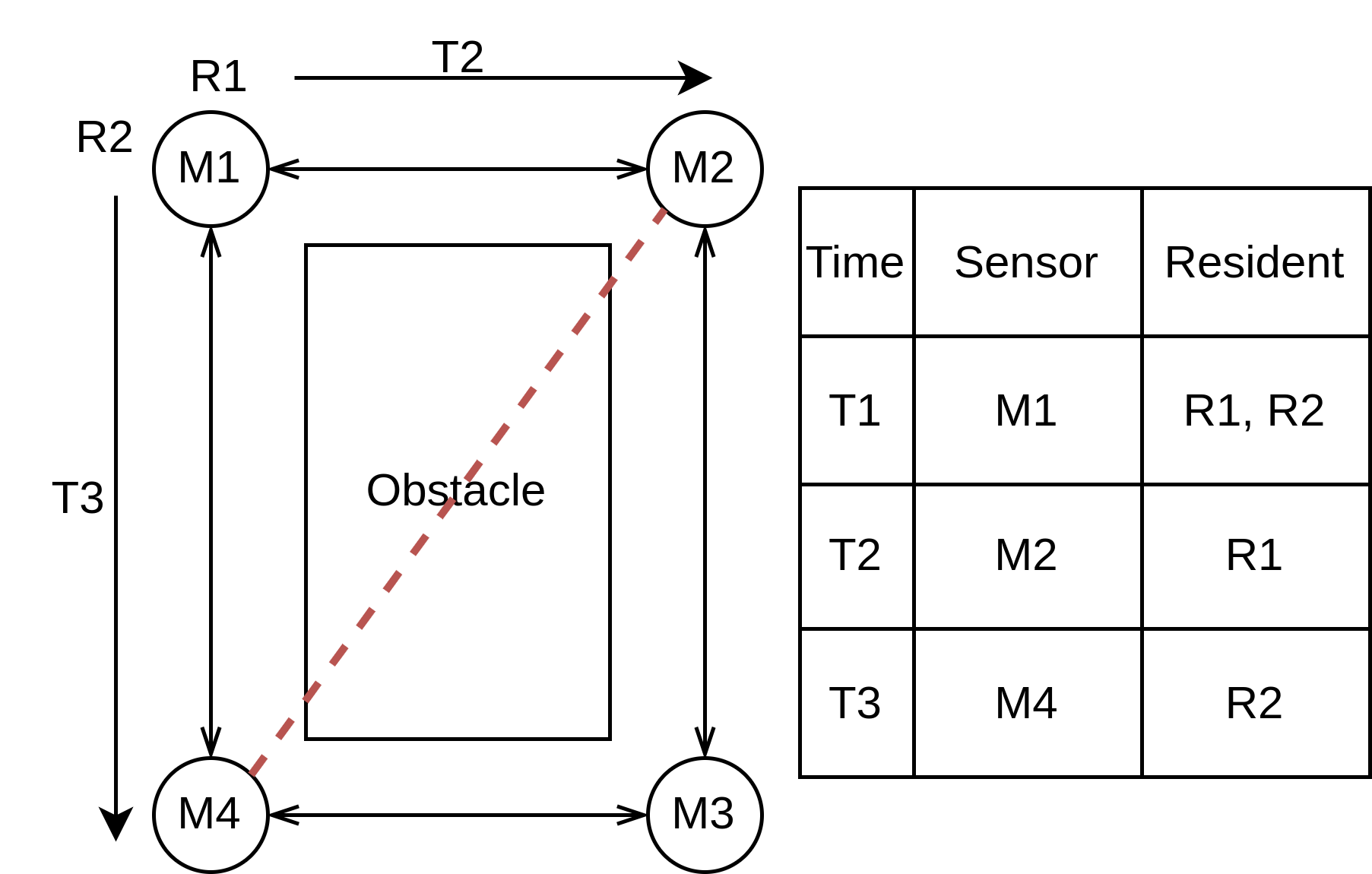}
        \label{fig:example1}
    }
    \hspace{2cm}
    \subfloat[Positional context of location.]{
        \includegraphics[width=4.2cm]{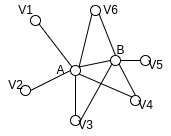}
        \label{fig:positional_context}
    }
    \caption{Illustration of topological context.}
\end{figure}

A prerequisite to ensure occupants' convenience is to \emph{identify residents} first and then provide seamless and personalized services to them. Therefore, this paper focuses on \emph{resident identification} in multi-resident smart homes. \emph{Resident identification is referred to as identifying occupants in smart homes}. The resident identification process uses residents' historical activities to determine which resident is responsible for triggering the sensor events. The sensor events are usually collected by sensors installed at different home locations. Their activities are usually sorted by time (i.e., temporal sequence of events). Resident identification in the home environment, however, is a difficult task due to the following challenges:

\begin{itemize}

\item The first challenge is the \emph{interaction effect} in the multi-occupant home. Decisions made by one resident may affect the decisions of another. For example, in a confined area, a resident's movement may block the movement of the other. This interaction is difficult to predict \cite{ChakiDipankar2020ACDF}. Meanwhile, when multiple occupants stay close to each other, interpreting the sensor events is complicated. In this scenario, a single sensor may detect multiple residents' activities in a short period of time. It is hard to distinguish which resident is the direct cause of each sensor event detected that sensor.

\item The second challenge is the \emph{lack of positional information} on the temporal sequence of events. The movement patterns of residents are often both temporal and spatial. It is easy to represent activity events either in a temporal sequence or spatial sequence separately. However, it is difficult to integrate them together. For a sequence only containing spatial sequence, the model might not be able to predict residents' activity based on their temporal habits. In contrast, with only temporal information included, it is hard for models to understand activities having spatial dependencies. Some events are \emph{unlikely to happen spatially} by multiple residents. For example, the left side of Fig. \ref{fig:example1} represents a room layout map. Four motion sensors (M1, M2, M3, M4) are installed at each corner, and an obstacle is at the center of the room. The right side of the figure is an annotated sensor event log. Considering two residents (R1 and R2) at the M1 position at time T1. At T2, R1 moves from M1 to M2, and at T3, R2 moves from M1 to M4. For a model without positional information knowledge, it is possible to consider the sequence $M1 \rightarrow M2 \rightarrow M4$ are triggered by the same resident. However, it is impossible spatially since residents must either pass M1 or M3 first before moving from M2 to M4 because of the center obstacle. A new way of encoding spatial information about the home to be included in the model might fill this gap.

\end{itemize}

We propose a location-aware framework for resident identification in multi-occupant smart home environments. It utilizes non-intrusive sensors (e.g., motion sensors and door sensors) as location indicators of residents \cite{10.1145/2528282.2528294}. This type of sensor is only triggered when activities happen nearby, making them perfect location indicators of \emph{entities}. We define entities as anyone that is movable by themselves. Of particular interest in this research is \emph{residents/occupants} in smart homes. Usually, a layout map with the precise location of obstacles and sensors is obtainable in a confined area \cite{Thomas2015}. Topological information could be extracted from those existing data. We consider fixed locations of sensors as a graph. This graph contains sensors' locations as vertices and residents' possible direct pathways between sensors as edges. Residents' movement is interpreted as movement inside the graph. The topological information of such a graph might be beneficial in the smart home environment since it may reflect the home's structure.

We consider locations as context-sensitive since they are dependent on each other. A movement to a location would not be possible without passing its neighbor. For instance, Fig. \ref{fig:positional_context} is an example of the context of positions. In this example, V1, V2, V3, V6, B are neighbors of A; A, V3, V4, V5, V6 are neighbors of B. The neighborhood (i.e., surroundings of nodes A and B) of center nodes represents the context of location. The contextual information reflects the connectivity and distance between locations. In the example, the center locations, A and B, depend on their connected neighbors. In order to have any entity move the center node A, it must pass one of the center node's neighboring nodes.

However, the naive representation of graphs is not suitable to be used in machine learning models. There is no simple way to directly integrate the naive graph representation into the temporal sequence of the event log. In order to make such graphs useful, the graph needs to be transformed into machine learning-friendly forms. The algorithm will transform the graph into a position encoding that retains the topological information of the original graph. \emph{Positional encoding is a way of encoding information about locations into finite-dimensional embeddings} \cite{rethinking_pos_encoding}. The encoding usually includes information about locations. Node embeddings are a type of positional encoding where a multi-dimensional vector represents each node of the graph. In the proposed framework, the constructed graphs are transformed into node embeddings. Vectors representing the node contain topological and connectivities information of the original graph. Context of locations is included in such positional encoding. The \emph{Node2Vec algorithm} is applied to help transform graphs into vector form so that a network of locations can be encoded \cite{node2vec}. The final output of Node2Vec is \emph{node embeddings}, where \emph{each sensor's location, connectivity, and topology are encoded into a vector representing each sensor}. The vector contains local connectivity and topological information about locations. When including such information in the sensor event log, we will be able to consider the sequence of events both temporally and spatially. After encoding, the node embeddings preserve such contextual information in the representation. We then concatenated it into the temporal sequences of sensor events and used it to train a \emph{Recurrent Neural Network (RNN)} classifier that identifies residents.


RNN models are implemented for resident identification tasks. The resident identification process requires a model making sequences of predictions on residents' identity using temporal sequences of activities as input. The process could be considered sequence-to-sequence mapping. \emph{The input sequences of events are mapped into an output sequence of residents' identities}, where the Seq2Seq model may be helpful. Especially the \emph{Long Short-Term Memory (LSTM)} is suitable for this task \cite{lstm}. LSTM is one of the proven Seq2Seq models and is widely used in other domains for Seq2Seq problems, such as time series analysis \cite{kim2019predicting}. LSTM was invented to resolve the major issue that traditional RNNs have, such as the \emph{vanishing gradient problem}, where the model tends to be unable to handle events with long dependencies. The main contributions of this work are:

\begin{itemize}
    \item A novel feature extraction technique that extracts high-dimensional, topological information from low-dimensional naive positional encoding. It is achieved by transforming layout maps into an accessibility graph and applying it to the Node2Vec algorithm.
    
    \item A novel resident identification model that employs Long Short-Term Memory (LSTM). The extracted positional encoding is integrated with the temporal sequences of sensor events as input of the LSTM model. To the best of our knowledge, it is the first attempt to integrate positional information with the temporal sequence of sensor events in resident identification.
    
    \item Design and execution of a smart environment setup and evaluated the proposed approach with real-world datasets. Experimental evaluation exhibits the efficiency and effectiveness of the proposed approach.
\end{itemize}

Section \ref{sec:related_works} discusses the existing works that use intrusive sensors and non-intrusive sensors. It also discusses existing positional encoding methods and Seq2Seq models in other domains. In section \ref{sec:preliminaries}, some concepts are formally defined for better explanation and understanding. In section \ref{sec:methodology}, the proposed method is explained in detail. After that, in section \ref{sec:experiment}, the smart environment setup and data collection process are introduced along with experiments. In section \ref{sec:result_discussion}, the effect of hyper-parameters of the proposed framework is evaluated, and the framework's performance is compared and discussed. Section \ref{sec:conclusion} concludes the paper with future works.

\section{Related Work}\label{sec:related_works}

We briefly introduce related works on (i) \emph{resident identification} in multi-occupant smart homes, (ii) \emph{positional encoding} technique and its possible applications motivating us to construct and utilize a graph to include topological information, and (iii) \emph{Seq2Seq models} used for converting sequences of events to the identification of the residents inspiring us to use Seq2Seq models to perform resident identification.


\subsection{Resident Identification}

Existing works on resident identification usually employ intrusive sensors such as cameras and microphones. Therefore, visual-based solutions are proposed for recognizing residents' identities \cite{kinect, 5733403}. These solutions process images by computer vision algorithms. Residents' body postures are extracted and used to identify residents \cite{kinect}. Grey-scale cameras are installed at multiple places to track residents' locations. Some resident identification systems exist based on voice recognition using \emph{microphones}. Residents are identified by their voice biometrics \cite{voice_biometric}. Another category of intrusive sensors is wearable devices such as smartwatches and smart bands. Bluetooth packets emitted by wearable devices are utilized to fingerprint the residents \cite{wearable_device_1}. Some other similar approaches use wearable tags for the identification of residents \cite{wearable_tags}. However, wearable devices are generally considered uncomfortable and inconvenient \cite{li2020your}.



Naive Bayes and Hidden Markov Models (HMM) are popular techniques that utilize temporal sequences of \emph{non-intrusive} sensor events. These statistical models trace users' activities for the task of resident identification \cite{Lesani2021, multi_resident_nb_hmm, hmm_resident_identification_2010, Baratchi_2014,nb_hmm2}. However, the Markov assumption addresses that the current state only depends on the previous state, while it is not true for events in smart environments. Events are usually recorded in sequences, and adjacent events have correlations. The current event not only depends on the current event but also on multiple previous events. Thus, models that could consider more historical events are more suitable for this task. Pattern mining algorithms are also developed to extract residents' activity traces from sensor logs \cite{ChenDongActivityRecognition}. The significant Correlation Pattern Miner (SCPM) algorithm is designed to extract usage patterns and correlations from appliance usage logs \cite{copminer}. SCPM is developed based on a generic pattern mining algorithm, PrefixSpan \cite{prefixspan}. It extracts usage patterns from temporal sequences of sensor activities that may identify residents in a multi-occupant environment.  A bag of events is a way to categorize events into activities that residents are currently performing. When considering all activities being performed in a short time interval, the pattern can be used to distinguish residents' identities during the time \cite{Mekuria2021}. The authors of this work consider events over a period of time as a way to detect residents' identities. This approach does not preserve The order of events, and the reported sensor logs are considered vectors. The vectors represent how many events are recorded by sensors in that period of time. All sensor events are labeled with residents' activities triggering the event. The authors believe there are patterns in the vector when different residents perform the same tasks. Resident identification is possible by using this pattern. A supervised Bayes Network is employed to identify residents. It requires annotation with the specific activities the residents are currently performing in addition to the resident's identity to work.


\subsection{Positional Encoding}

\emph{The positional encoding technique allows machine learning models to focus on specific data subsets instead of the entire dataset.} The positional encoding can be used to encode positional or topological information originally in formats that are not useful in machine learning algorithms into vector forms. The specific subset of data may reveal a strong correlation with the intended task of the model \cite{yun2019transformers}. Positional encoding techniques used in other domains (e.g., natural language processing (NLP) and image processing) are inspiring us for the resident identification problem. For example, in NLP domain, the transformer model uses positional encoding to associate the position of words in the original sequence of the sentence \cite{transformer}. For the transformer model, the words are processed simultaneously. The positions of words in sentences are unknown to the transformer model. Positional encoding inserts the positional information of the word into the sentence. The positional encoding might be used as biases to boost the performance of a Generative Adversarial Network (GAN) \cite{Xu_2021_CVPR}. The GAN is used for image processing where incorporating positional encoding increases the weight of important areas of images. Multiple positional encoding techniques exist that transform graphs into positional embeddings \cite{graph_contrastive_coding,dwivedi2020benchmarkinggnn,jin2020gralsp}. These methods usually mine out structural patterns from graphs and convert vertices, edges, or subgraphs into low-dimensional embeddings. These patterns are then applied to standard machine learning models. In smart environments, positional encoding techniques usually extract locations of interests and structures into embeddings. The embeddings can be concatenated to input parameters, such as sensor events arranged in temporal sequences, to provide location awareness to the sequence. A similar technique could be used to address the resident identification problem, much like the application of positional encoding mentioned above. Twomey, N. et al. proposed a way of learning topologies from sensor event sequences into graphs \cite{twomey2017unsupervised}. However, they did not consider integrating temporal information with the learned positional information. \emph{Those ideas, which convert graphs into vector forms, motivate us to use the Node2Vec algorithm to transform the graph into node embeddings that are easily integrated into machine learning models.}


\subsection{Seq2Seq Models}

The wide usage of Seq2Seq models in other domains motivates us to adopt them for resident identification problems. We use Seq2Seq models as a tool to build the resident identification model. Seq2Seq models usually have an architecture that consists of an encoder and a decoder. The encoder encodes the input into intermediate hidden states, and the decoder decodes the hidden states into desired output form \cite{cho2014learning}. RNN is widely used for this type in such architecture. LSTM and Gated Recurrent Unit (GRU) are two popular RNN cells \cite{DBLP:journals/corr/ChoMBB14}. On the one hand, LSTM can remember long-term patterns. Besides, it also can forget events that occurred too long ago and are not revised by the model. On the other hand, GRU can be considered a simplified version of LSTM since it does not have the cell state output that LSTM has. It has fewer training parameters than LSTM. However, due to the smaller parameter space, its performance on larger datasets or complex problems may be outperformed by LSTM. LSTM is widely used to analyze time series data, such as predicting power fluctuations and stock prices \cite{8039509, 8126078, siami2018comparison}, predicting driver's identity \cite{10.1145/3412353}. In the smart home domain, LSTM is used to extract residents' activities from various data sources \cite{liciotti2020sequential, mutegeki2020cnn}. \emph{The encoder and decoder architecture allows Seq2Seq models to map input data into output vectors.} When considering the Seq2Seq models, the sequences of events generated by residents can be mapped into residents' identities by the model. \emph{The benefits of Seq2Seq models inspired us to use them in resident identification problems.} The resident identification problems shared similarities to other problems where Seq2Seq models could be applied.

In summary, the positional encoding technique could also be applied to the resident identification problem. Resident identification models may know more about residents' activity patterns when positional information is included. Meanwhile, intrusive sensors such as cameras and wearable devices are widely used in the existing literature. However, they are sensitive for installation in home environments \cite{li2020your}. Besides, it is uncomfortable and inconvenient for residents to have wearable devices always carried on. Furthermore, visual or sound-based methods are usually considered to be computationally expensive. Existing works for resident identification, including Naive Bayes and Hidden Markov Chain, usually only analyze temporal sequences of events. Positional information about the home is ignored. In this work, non-intrusive sensors, such as motion and door sensors, are used. These non-intrusive sensors do not have direct interference with residents. To the best of our knowledge, no previous work has been done on integrating positional information with temporal sequences. \emph{Therefore, this research emphasizes the importance of positional encoding of smart homes and their integration with the temporal sequence of events. The added topological information about the smart environment increases the performance of the resident identification model.}



\section{Preliminaries}\label{sec:preliminaries}

Positional encoding is defined as a way to encode positional information \cite{rethinking_pos_encoding}. In this context, the term position refers to the sensors' locations. In a smart environment, sensors are usually used for monitoring entities' movement. Entities are anything that is capable of moving by themselves — for example, residents, pets, or robots. Sensors like door and motion sensors will only get triggered by nearby activities of entities. In this research, we only focus on one type of entity: the resident. Residents' activities are both temporal and spatial\cite{Huang_2018}. In addition to temporal patterns during their activities, activities usually have spatial movement patterns.

\begin{figure}[!t]
    \centering
    \includegraphics[width=6.5cm]{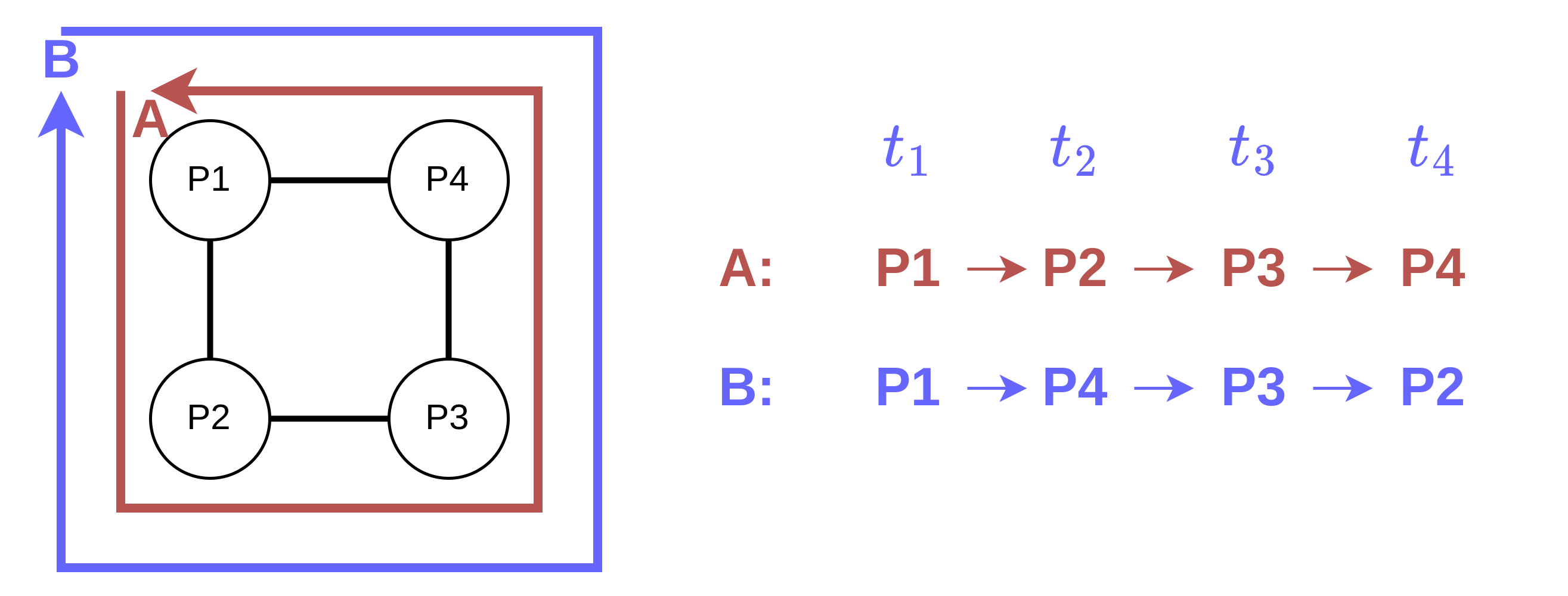}
    \caption{Two residents moving in two opposite patterns.}
    \label{fig:example2}
\end{figure}

For example, a resident studies in the study room and decides to go for lunch. The door between the study room and corridor and the door between the corridor and dining hall must be passed by this particular resident. In this example, the system may recognize a resident's activities: stay stationary at study, move to the dining hall, pass through the corridor, and keep stationary at the dining hall. If this sequence of events happens repeatedly, the system may associate future sensor events with similarity being triggered by the same resident in sequences. For instance, it may predict that the resident is more likely to transition from working in the study to having dinner in the dining hall when observing such events. The sequence of locations of a resident's movement can be used for the analysis of the activities of residents. This information might be helpful for resident identification problems. The added context awareness also allows for distinguishing between similar patterns. The following example (Fig. \ref{fig:example2}) consists of 4 POIs (P1, P2, P3, P4) forming a rectangle. There are 2 entities/residents (A and B). In this context, entity A moves clockwise, whereas entity B moves anticlockwise. The model performs recognition considering the residents' movement patterns since the order of the POI passed differs between the two entities. It is difficult for models to distinguish differences in a system with only the knowledge of each event's timestamp. However, with positional encoding, the context of positions could help models distinguish the difference.

One of the core contributions of this research is a novel feature extraction technique that extracts high-dimensional, topological information from low-dimensional naive positional encoding, typically a set of coordinates. We consider positional encoding of Point of Interest (POI) in a vector form. GPS-encoded coordinates usually represent a POI in a 3-tuple of latitude, longitude, and altitude, forming a vector of size $1\times 3$. This type of positional encoding is considered low-dimensional positional embeddings. It may have limited information about connectivities between nodes. In this regard, the proposed technique transforms such low-dimension positional embeddings into high-dimensional embeddings. These high-dimensional embeddings include topological and contextual information about POIs and knowledge about POIs' connectivity. The feature extraction technique first builds an accessibility graph with knowledge about the coordinates of POIs and obstacles. Then, the graph is transformed into an APG that finally trains the Node2Vec node embeddings (definition \ref{def:node_embeddings}). Finally, the generated node embeddings are concatenated into temporal sequences of user activities. This adds spatial contextual information to the original data. It allows further machine learning models to better understand residents' activity patterns by integrating both temporal and spatial information.

\begin{definition}\emph{Entity.}
\label{def:entity}
An entity is any object usually able to move by itself in the home environment. Entities are not necessarily limited to residents. Pets or robots can also be regarded as entities. Any stationary object may not be considered as an entity. For example, furniture and appliances are seldom moved, and their movements can only happen when assisted by entities.
\end{definition}

\begin{definition}\emph{Point of interest (POI).}
\label{def:poi}
A point of interest refers to a location where entities could interact with, present at, or an event could happen. Those are locations where this work is trying to encode. They are generally accessible by residents. For example, it could be a junction of corridors or appliances like refrigerators.
\end{definition}


\begin{figure}
    \centering
     \subfloat[Original.]{
        \includegraphics[width=4cm]{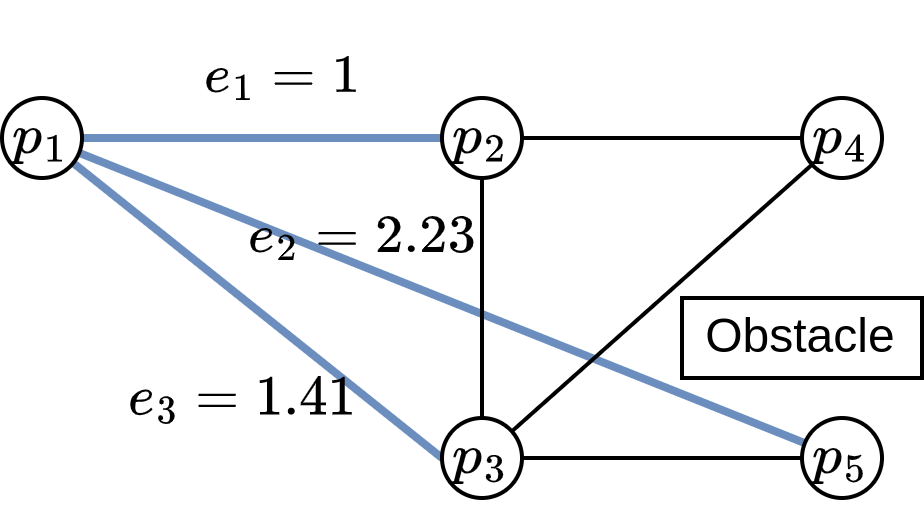}
        \label{fig:acc_graph}
    }
    \subfloat[Probability.]{
        \includegraphics[width=4cm]{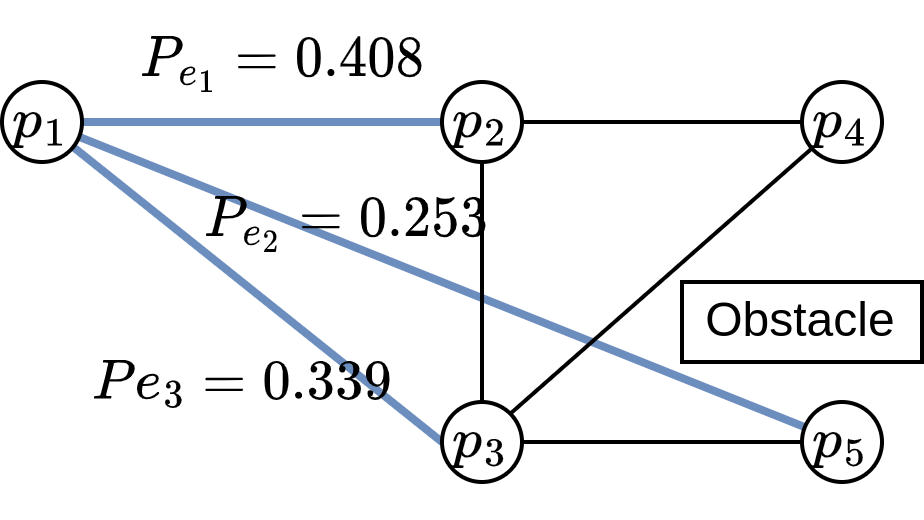}
        \label{fig:acc_prob_graph}
    }
    \caption{Accessibility graph and accessibility probability graph.}
    \label{fig:acc_and_acc_prob_graph}
\end{figure}

\begin{definition}\emph{Accessibility graph (AG).}
\label{def:acc_graph}
An accessibility graph $G = \{P, E\}$ is a connected, weighted, undirected graph, of which,
vertices $P = \{p_1, p_2, \cdots, p_n\}$ being POI.
Edges $E = \{e_1, e_2, \cdots, e_m\}$ are connecting 2 POIs representing the only straight possible pathway, without any obstacles and other POIs, on the edge route.
The edges are weighted by the distance between 2 POIs.
\end{definition}

\begin{definition}\emph{Accessibility probability graph (APG).}
\label{def:acc_prob_graph}
An accessibility probability graph is similar to the accessibility graph. However, instead of weighing the edges based on the distance between 2 POIs, it weighs the edges by probabilities of an entity moving between the two ends of the edge.
\end{definition}

Fig. \ref{fig:acc_graph} is a simple example of an AG. It does not contain any edges cut by obstacles. The figure shows that $p_4$ and $p_5$ are not connected due to the obstacle in between. Fig. \ref{fig:acc_prob_graph} is an example of an accessibility probability graph, and typically, the APG could be transformed from a corresponding AG by Algorithm \ref{alg:t_mat_from_adj_mat}. The edges of APGs are weighted by the probability of entities transiting between nodes. For example, $P_{e_1}$ is the probability for entity transit from POI $p_1$ to $p_2$. And it can be calculated by $$P_{e1} = \frac{\frac{1}{e_1+1}}{\frac{1}{e_1 + 1} + \frac{1}{e_2 + 1} + \frac{1}{e_3 + 1}}= 0.408 $$

\begin{definition}\emph{Node embeddings.}
\label{def:node_embeddings}
Node embeddings are a type of positional encoding that encodes graphs into vector form. It is similar to the concept of word vector embeddings (i.e., using the vector of numbers trained by some methods to represent the word). The trained embeddings include information about the word learned from the corpus. Typically, such information is extracted from the surrounding words of the central word. We can understand it as using contextual information of words in a corpus to represent the word in a vector form. The same idea applies to the graph as well. The graph's nodes could be represented in vector form by using similar algorithms that use the contextual information of the graph to represent the node.
\end{definition}

\section{Methodology}\label{sec:methodology}

\begin{figure}[!t]
    \centering
    \includegraphics[width=8.5cm]{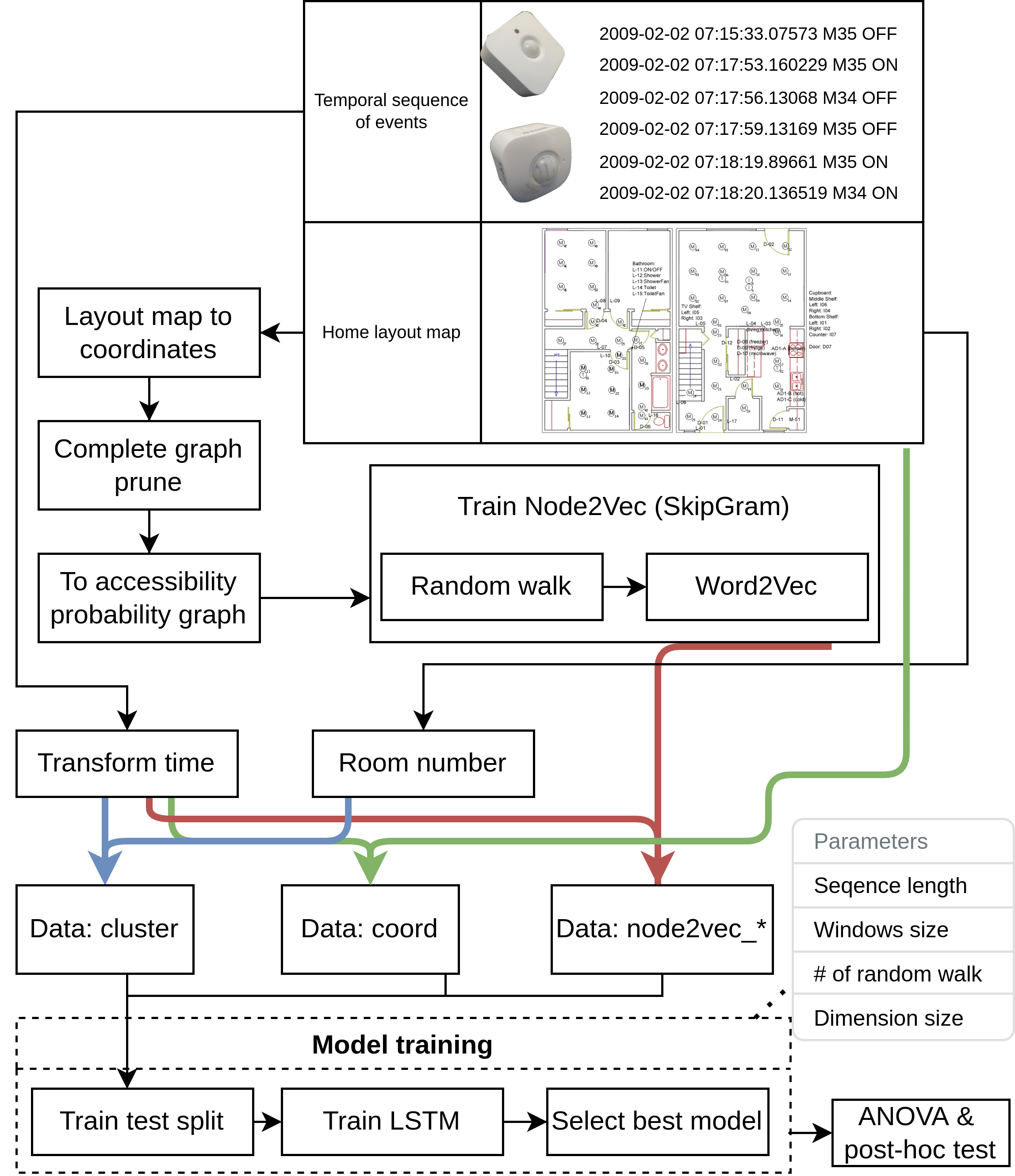}
    \caption{The architecture of the proposed framework.}
    \label{fig:system_arch}
\end{figure}

The general architecture of the proposed framework is shown in Fig. \ref{fig:system_arch}. The framework accepts two data components as input. An annotated temporal sequence of sensor event logs (i.e., the temporal sequence of events) and coordinates of installed sensors and obstacles in the home environments (i.e., home layout map). 

At first, time components of timestamps, namely day-of-year, weekday, and second-of-day, are transformed into a $1\times 2$ vector by Equation (\ref{eq:time_transformation}). This converts the linear representation of time into a chronic form. Then, the layout maps are transformed into a machine learning-friendly form. Coordinates of sensors and obstacles in the smart environment are transformed into an AG (Definition \ref{def:acc_graph}) by Algorithm \ref{alg:cgp}. Then it has been further transformed into an APG (Definition \ref{def:acc_prob_graph}) by Algorithm \ref{alg:t_mat_from_adj_mat}. The access probability graph is used as input for the Node2Vec algorithm, where nodes of the graph are further transformed into node embeddings. The Node2Vec embeddings contain topological information, including the connectivity between nodes and their relative distance into the final output form. 
Finally, a supervised LSTM model is trained, validated, and evaluated with data generated from previous steps. The data being used for training is a temporal sequence of sensor events with extra parameters, which are node embeddings. They add positional information and structural information about the home into the model.



\subsection{Transformation of Time}



\begin{figure}[t]
    \centering
        \includegraphics[width=8.5cm]{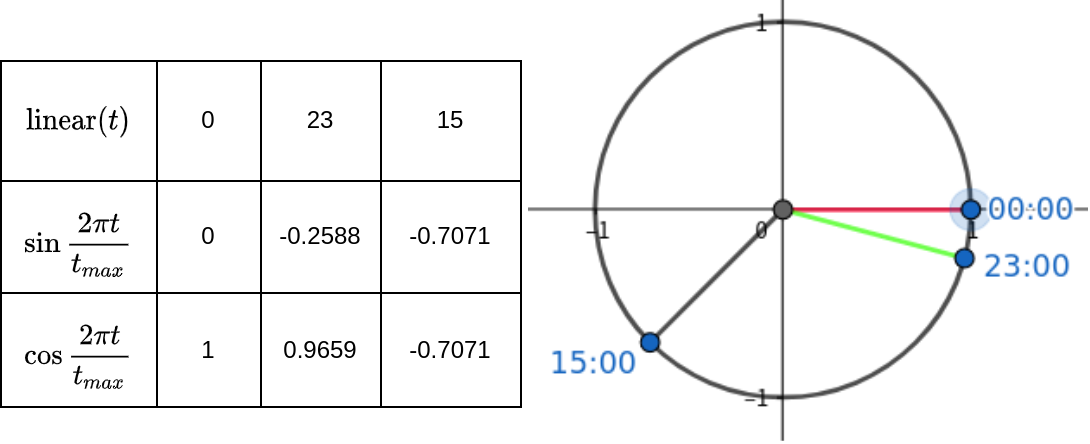}
        \captionof{figure}{Example of the time transformation.}
        \label{fig:time_trans_example}
\end{figure}

The time transformation is applied as a data pre-processing step. The transformation converts the timestamp in the scalar into a $1 \times 2$ vector. The vector is chronically compared to its linear form. The reason for such conversion is that it is usually chronically against time when considering users' habits. Residents' activities tend to have correlations with time points of days. However, the naive linear way of encoding time in timestamp does not preserve such property. Considering the two dates, January 1 and December 31, they are very different if encoded in a linear timestamp. At the same time, residents' activity patterns may still share great similarities. For instance, they would significantly differ if we encode the two dates into seconds since epoch time (i.e., 1970-01-01). The value between the two dates will have a 31536000-second difference. In this regard, we use a triangle transformation on time components (i.e., days and seconds) to convert them into vectors with two elements. Equation (\ref{eq:time_transformation}) is used for such transformation \cite{time_encoding}. Fig. \ref{fig:time_trans_example} shows the chronic property of encoded time after transformation.


\begin{equation}
t_{\sin} = \sin{\frac{2\pi t}{t_{max}}},
\hspace{2cm}
t_{\cos} = \cos{\frac{2\pi t}{t_{max}}}
\label{eq:time_transformation}
\end{equation}

In Equation (\ref{eq:time_transformation}), $t$ is an integer representing a time component started with 0, and $t_{max}$ is the max possible value of the time component. For example, if we are interested in hours as one of the time components, consider three hours of a day $h_1 = \text{00:00}$, $h_2 = \text{23:00}$ and $h_3 = \text{15:00}$. Since there are a maximum of 24 hours per day, we have $t_{max} = 24$. $h_1$ is the first hour-of-day, $t_1 = 0$; $h_2$ is the last hour-of-day, $t_2 = 23$; $h_3$ is the 15th hour-of-day, $t_3 = 15$. This example is visualized in Fig. \ref{fig:time_trans_example}.
Cosine distance is used to calculate the distance between two components of time. For example, the distance between $t_1$ and $t_2$ is 0.03407 while the distance between $t_1$ and $t_3$ is 1.7071. However, if we encode these three hours of a day linearly, the distance between $h_1$ and $h_3$ is smaller than the distance between $h_1$ and $h_2$.

We select and encode three components of time by this method. They are (i) day-of-year, (ii) day-of-week, and (iii) second-of-day. Those three components are chosen because residents usually have chronic living patterns within those time frames. The seasonal effects on residents' habits will be reflected by the day of the year since there is a chronic pattern in different years on the same day of the year. Another example is that workdays and weekends may have different living habits, i.e., the resident may wake up late on weekends. This workday/weekend variable may reflect these patterns. To apply the algorithm, take the timestamp \verb|Thursday 2023-08-24T15:00:00| as an example. We firstly separate it into 3-time components: (i) 24th August, the 236th day of the year, and there is a total of 365 days; (ii) Thursday, the 3rd day of the week; (iii) 15:00:00, the 54000th seconds of the day. We can then apply Equation (\ref{eq:time_transformation}) to transform these time components. \footnote{We count starting from 0 and consider Monday as the first day of a week} Later, those vectors will be used as replacements for timestamps to include the chronic property of time to model.

\begin{equation*}
T_{\text{day}} = \left\{\begin{aligned}
    t_{sin} &= \frac{236 \times 2\pi}{365} = -0.796 \\
    t_{cos} &= \frac{236 \times 2\pi}{365} = -0.605 \\
\end{aligned}\right.,
T_{\text{weekday}} = \left\{\begin{aligned}
    t_{sin} &= \frac{3 \times 2\pi}{7} = 0.434 \\
    t_{cos} &= \frac{3 \times 2\pi}{7} = -0.901\\
\end{aligned}\right.,
T_{\text{second}} = \left\{\begin{aligned}
    t_{sin} &= -0.707 \\
    t_{cos} &= -0.707 \\
\end{aligned}\right.
\end{equation*}

\subsection{Building Accessibility Graph from Coordinates}\label{sec:cgp}

We design an algorithm (Algorithm \ref{alg:cgp}) to help the construction of AGs from coordinates of POIs and coordinates of obstacles. We name the algorithm \emph{complete graph prune}. As its name suggests, the algorithm constructs the AG by building a fully connected graph of all POIs. Vertices of the graph are within the same coordinates system and have their coordinates labeled. The algorithm considers obstacles as two coordinates forming a line segment, being in the same coordinates system with the nodes of the fully connected graph. After that, a line segment collision detection method is applied. The line segment collisions are applied to all edges over all obstacles. Once a collision has been detected, the edge connecting the two nodes is removed.



The line segment detection function used in the algorithm is at constant time complexity. It can handle line segments consisting of multi-dimensional coordinates. \emph{The two line segments $L_1 = (P_1, P_2)$ and $L_2 = (P_3, P_4)$ are collided if and only if the $P_1$ and $P_2$ are on the different side of $L_2$, and $P_3$ and $P_4$ are on the different side of $L_1$}. 
We can use cross-products of subtraction between coordinates to detect if two points of a line segment are on different sides of another line segment. If among the four points of the two lines, no three points are on the same line, we only need to make sure the above property holds for the four points. The cross-product taken between the four points is used to determine if the four points are all on both sides of each other. Each of the cross products determines if the given point is on the right-hand side of the line segment. For example, $v_1 = (C_1 - C_3) \times (C_4 - C_3)$ indicates $C_1$ is on the right-hand side of line segment $L_2 = (C_3, C_4)$ when $v_1 < 0$. Since we need to make sure the two points of $L_1$ on both sides of $L_2$, the sign of $v_1$ and $v_2$ should be different. Thus $v_1 \dot v_2 < 0$ is used to perform this check. When one of the points is not on the side of the line segment, it must be in the line holding the segment. In this scenario, we need to determine if the point is in the middle of the line segment. The \verb|OnSegment| is doing this check; it checks if $P$ is in the middle of line segment $L$. The pseudo-code of the proposed algorithm for building the AG, including the line segment detection function, is defined in Algorithm \ref{alg:cgp}.

\begin{algorithm}[t]
\caption{Complete graph prune}\label{alg:cgp}
\begin{algorithmic}[1]

\Input
    \Desc{$C$}{$= \{c_1, c_2, \dots, c_n\}$, A set of POI in coordinates}
    \Desc{$W$}{$= \{(w_{1a}, w_{1b}), (w_{2a}, w_{2b}), \dots, (w_{ma}, w_{mb})\}$, A set of obstacles in 2-tuple of coordinates as line segments}
\EndInput
\Output
    \Desc{$g$,}{An accessibility graph}
\EndOutput
\State Let $g = (C, E), E = C \times C$
\For{$e \in E$}
    \For{$w \in W$}
        \If{$\text{LINESEGDETECT}(e, w)$}
            $E \backslash \{e\}$
        \EndIf
    \EndFor
\EndFor

\Function{LinesegDetect}{$L_1 = (C_1, C_2), L_2 = (C_3, C_4)$}
    \State $v_1 = (C_1 - C_3) \times (C_4 - C_3)$
    \State $v_2 = (C_2 - C_3) \times (C_4 - C_3)$
    \State $v_3 = (C_3 - C_1) \times (C_2 - C_1)$
    \State $v_4 = (C_4 - C_1) \times (C_2 - C_1)$
    \If{($v_1 \cdot v_2 < 0) \land (v_3 \cdot v_4 < 0$)}
        \State \Return True
    \ElsIf{ $v_1 = 0 \land \text{onSegment}(C_3, C_4, C_1)$ }
        \State \Return True
    \ElsIf{ $v_2 = 0 \land \text{onSegment}(C_3, C_4, C_2)$ }
        \State \Return True
    \ElsIf{ $v_3 = 0 \land \text{onSegment}(C_1, C_2, C_3)$ }
        \State \Return True
    \ElsIf{ $v_4 = 0 \land \text{onSegment}(C_1, C_2, C_4)$ }
        \State \Return True
    \EndIf
    \State \Return False
\EndFunction

\Function{onSegment}{$L_1 = (P_1, P_2)$, $P$}
    \For{$i \in 0..Dimension(P_1)$}
        \If{$\neg \min{(P_{1i}, P_{2i})} < P_i < \max{(P_{1i}, P_{2i}}$)}
            \State \Return False
        \EndIf
    \EndFor
    \State \Return True
\EndFunction
    
\end{algorithmic}
\end{algorithm}

\begin{figure}[t]
    \centering

    \subfloat[Input.]{
        \includegraphics[width=4.5cm]{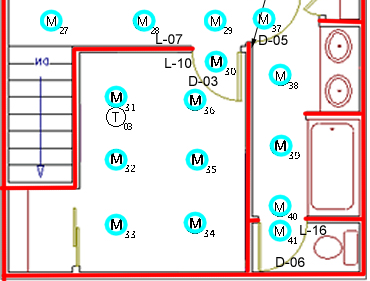}
        \label{fig:cgp_base}
    }
    \subfloat[Fully connected.]{
        \includegraphics[width=4.5cm]{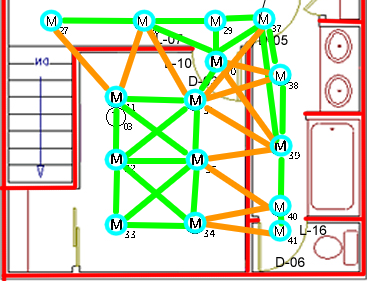}
        \label{fig:cgp_prune}
    }
    \subfloat[Pruned.]{
        \includegraphics[width=4.5cm]{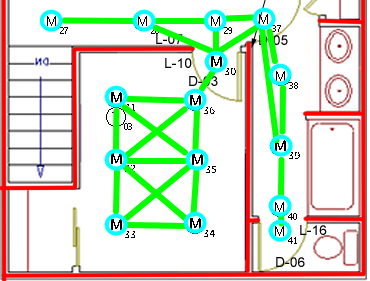}
        \label{fig:cgp_done}
    }
    
    
    \caption[cap]{Complete graph prune process \footnotemark.}\label{fig:cgp_process}
\end{figure}

Fig. \ref{fig:cgp_process} shows three intermediate states of Algorithm \ref{alg:cgp}. Fig. \ref{fig:cgp_base} shows the initial obstacles in red and the motion sensors in green. Fig. \ref{fig:cgp_prune} shows a complete graph of edges connecting all motion sensors, with green edges being the result and orange edges being the pruned edges due to collision with obstacles. The last figure (Fig. \ref{fig:cgp_done}) shows the result of the algorithm. We can find that the algorithm prunes all edges that collide with obstacles.

The algorithm uses a naive approach that requires a nested loop of all edges of a fully connected graph and all obstacles. Given there are $n$ POIs and $m$ obstacles represented by line segments, and the line segments collision detection algorithm used has a constant time complexity of $O(1)$, the time complexity for this algorithm is $O(\frac{n(n-1)m}{2}) = O(mn^2)$. Optimization of this algorithm is possible by splitting the graph into small partitions. If the graph is handled separately in $f$ partitions, and all partitions have a similar number of POIs and obstacles. That is, POIs for each partition are close to $\frac{n}{f}$, and the number of obstacles is close to $\frac{m}{f}$. The time complexity of the algorithm would be $O(f * \frac{\frac{mn}{f^2}(\frac{n}{f} - 1)}{2}) = O(\frac{mn^2}{f^2})$ if this optimization technique is applied. Therefore, if groups of large chunks of sensors have few connections in between, it is suggested to apply the algorithm individually and then manually add the connection between groups. This could reduce the amount of time required by the algorithm.

\footnotetext{Some edges are omitted for the aesthetic purpose}

\subsection{Accessibility Graph to Accessibility Probability Graph}

The built accessibility graph cannot be directly applied to the Node2Vec algorithm. The reason is that the edges of accessibility graphs are weighted by distance. In this case, the random walk process of the Node2Vec algorithm makes the close neighbors of sensors less likely to be visited. The random walk process is biased on distant nodes because of their higher weight. This is different for scenarios where residents move within homes. Therefore, the original accessibility graph needs to be transformed into an APG. Edges are weighted by the probability of transition between nodes instead of the absolute distance. In this work, We assume the average normal movement speed of residents is similar. This assumption is made considering the following scenario in Fig. \ref{fig:positional_context}. Considering there are two residents $R_1$ at $A$ and $R_2$ at $B$, and there are two sensor events $e_a$, $e_b$ both at time point $t_0$ reporting their presence at those two POIs. if there is a new event being recorded at $t_1$ at POI $V4$, since we are assuming the residents' movement speeds are similar, it is more likely that $R_2$ is making such movement. This assumption is made since it is not convenient to collect residents' real-life transition probability between POIs in smart environments; the resident identification model using the APG construct based on this assumption could observe satisfactory improvements; and the assumption is only relevant when used to transform the AG into APG, it is not used in the following resident identification models.



%


\begin{algorithm}
    \caption{Transition matrix from adjacent matrix}\label{alg:t_mat_from_adj_mat}
        \begin{algorithmic}[1]
        \Input
            \Desc{$M_g$,}{   adjacent matrix of a $n$-vertices $AG(V, E)$}
            \Desc{$w$,}{   diagonal weight}
        \EndInput
        \Output
            \Desc{$M_{\text{out}}$,}{   A adjacent matrix of an APG}
        \EndOutput
        \State Calculate the matrix mask of $M$
        \State For each entry $m_{ij}$ greater than 0 of matrix $M$, apply inverse function.
        \For{$i \in 1..n$}
            \For{$j \in 1..k$}
                \If{$m_{ij} > 0$}
                    $m_{ij} = \frac{1}{m_{ij} + 1}$
                \EndIf
            \EndFor
        \EndFor
        \State Normalize by row. $M = \frac{M}{\sum_{j=1}^{n}m_{ij}}$
        \State Add diagonals with weight. $M = M + \frac{wI}{1-w}$
        \State Normalize by row again. $M_{\text{out}} = \frac{M}{\sum_{j=1}^{n}m_{ij}}$
    \end{algorithmic}
\end{algorithm}


To construct APG from AG, a mathematical function is applied to transform the distance into a probability with the above assumption. The accessibility graph, $AG(V, E)$, is represented as an adjacent matrix, $M_g$. This $M_g$ is the input of the Algorithm \ref{alg:t_mat_from_adj_mat}. We then calculate the transition probabilities between locations. We use inverse function $f(M) = \frac{1}{M + 1}$ to calculate the weight because of the assumption mentioned above: the resident who is closer to the location where the last event occurred is more likely to be accounted for the new event. Intuitively, this function makes distant locations less likely to be visited, while nearer locations are more likely to be visited. The inverse function is applied to all matrix entries with a value greater than 0. The function is not applied to 0-value entries since the probability of transiting between two disconnected locations is 0. After calculating the probability weight, we normalize the matrix into a transition matrix so that the summation of a row becomes 1. After normalization, we can obtain an APG with all its diagonals being 0, as in the following example (i.e., $M_g$ matrix). It is also a transition matrix of a Markov chain. However, this APG requires further adjustments to fit into our model.

\begin{equation*}
    M_g = \left[
    \begin{array}{cccc}
        0 & 1 & 0 & 0 \\
        0.46 & 0 & 0.31 & 0.23 \\
        0 & 0.4 & 0 & 0.6 \\
        0 & 0.33 & 0.67 & 0 \\
    \end{array}
    \right]
\end{equation*}

In a home environment, residents may stay in a position for a long period of time. For example, sitting on a sofa to watch TV. During this period, their location will be fixed. However, by the current APG, the probability of staying stationary at a location (i.e., the self-loop probability) is zero. This scenario differs from other common scenarios where residents move within the home environment. When we are constructing the AG from the layout map, the diagonal of the matrix is considered to be 0 since the distance between the point and itself is non-existence. Therefore, we need to adjust the diagonal weight of APG. This adjustment is necessary since it is possible for residents to be detected and reported by the same sensor for a sequence of continuous events. Without the weight adjustment, the topology generated by the APG will consider a resident to be stationary, being impossible over time. To perform such adjustment, a diagonal matrix of $\frac{wI}{1-w}$ is added to the calculated matrix to counter this issue. It makes the self-loop probability into $w$. The last step normalizes the matrix again due to the added weight on diagonals. The following example shows the algorithm's input and output states, where $M_g$ is a $4\times 4$ AG as input; $M_{\text{out}}$ is the output APG. The output APG uses a diagonal weight of $w = 0.5$.

\begin{equation}\label{eq:example-apg}
\begin{aligned}
    M_g = \left[
    \begin{array}{cccc}
        0 & 1 & 0 & 0 \\
        1 & 0 & 2 & 3 \\
        0 & 2 & 0 & 1 \\
        0 & 3 & 1 & 0 \\
    \end{array}
    \right]
    M_{\text{out}} =
    \left[
    \begin{array}{cccc}
        0.5 & 0.5 & 0 & 0 \\
        0.23 & 0.5 & 0.15 & 0.12 \\
        0 & 0.2 & 0.5 & 0.3 \\
        0 & 0.17 & 0.33 & 0.5 \\
    \end{array}
    \right]
\end{aligned}
\end{equation}

This algorithm produces a transition probability matrix of a finite state Markov chain, with each state being the POI of the home environment. The probability of the resident moving from one POI to any other POI is represented by the transition probabilities of the matrix. This transition probability matrix could also be considered as an adjacent matrix of an APG, where it is essential for the next step to train the Node2Vec node embeddings.

\subsection{Transforming APG into Node Embeddings}


In this step, the Node2Vec algorithm introduced by Aditya and Jure is used to transform accessibility probability graphs into node embeddings \cite{node2vec}. It is a model that wraps the SkipGram model, which is usually used for generating word embeddings\cite{41224}. The method requires random walks on the given graph to generate sequences of vertices. The random walk process will at least be applied to all vertices as the beginning vertex. The trained APG will be used to perform the random walks. For example, if we consider the simple APG in Equation (\ref{eq:example-apg}). If we are performing a random walk process starting from the first node, we can get the probability of the next node by looking up the first row of the APG matrix. The next node would have 50\% probability being node 1 or 50\% being node 2. A random process will be applied to determine which node is to be selected using the given weight. The random walk process will be repeated several times until the sequence length limit is met. After the random walk process, the algorithm adopts the concept from the Word2Vec algorithm \cite{node2vec}. The sequences of vertices are treated as sentences, and vertices are considered words for the Word2Vec models. It then trains graph nodes as vocabulary into word embeddings. The underlying SkipGram model is a Word2Vec algorithm attempting to use the center word to learn the information of its contextual words. For node embeddings, like word embeddings, where the trained vector representation of words contains the contextual information about words in a document, the trained vector representation of nodes in a graph contains contextual information about its neighborhood nodes. The list of node sequences by the random walk process is then applied to the underlying SkipGram model of Node2Vec. After that, the model outputs node embeddings, with each node having an associated vector representing it. The trained vector includes contextual and topological information about the node's position and connectivity in the original graph. Such node embeddings are easily added to the temporal sequences of sensor activity data. The node embeddings are later concatenated as extra features to the temporal sensor events logs.


\subsection{The LSTM Model}

\begin{figure}[t]
    \centering
    \includegraphics[width=8cm]{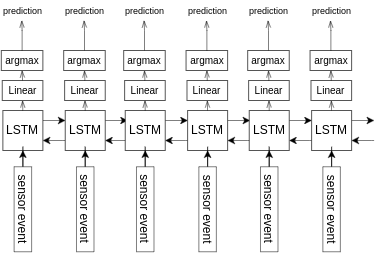}
    \caption{LSTM structure of the resident identification model.}
    \label{fig:lstm}
\end{figure}

A Seq2Seq, LSTM tagger is used for identifying residents. The model consists of a one-layer, bidirectional LSTM layer, an optional dropout layer, and a linear layer to transform the LSTM state into a tag space of $1 \times n$ vector, where $n$ is the number of residents to identify. The dropout layer is helpful when a class imbalance exists. The structure of the LSTM model is presented in Fig. \ref{fig:lstm}. An LSTM unit consists of multiple memory cells that could hold temporal and spatial information about residents' activity patterns. Within the memory cell, there are two special activation functions, namely sigmoid ($\sigma$) and tanh function as defined in Equations (\ref{eq:sigmoid}) and (\ref{eq:activation}).

\begin{equation}
    \sigma(x) = \frac{1}{1+e^{-x}}
    \label{eq:sigmoid}
\end{equation}
\begin{equation}
    tanh(x) = \frac{e^x - e^{-x}}{e^x + e^{-x}}
    \label{eq:activation}
\end{equation} 

It achieves a short-time memory by a series of gates in its recurrent units. Three gate units are defined for each LSTM unit. They are the input gate ($i_t$), forget gate ($f_t$), and output gate ($o_t$). Those three gates utilize the sigmoid activation function to enable their ability to bias the weight into 0 as represented in Equation (\ref{eq:lstm_gates}).

\begin{equation}
\begin{aligned}
i_t &= \sigma(w_i[h_{t-1}, x_t] + b_i) \\
f_t &= \sigma(w_f[h_{t-1}, x_t] + b_f) \\
o_t &= \sigma(w_o[h_{t-1}, x_t] + b_o) \\
\end{aligned}
\label{eq:lstm_gates}
\end{equation}

The following equation calculates the cell output:

\begin{equation}
\begin{aligned}
\tilde{c_t} &= tanh(w_c[h_{t-1}, x_t] + b_c) \\
c_t &= f_t c_{t-1} + i_t \tilde{c_t} \\
h_t &= o_t tanh(c^t) \\
\end{aligned}
\end{equation}

The cell states and the hidden states are fed into the following cells at timestamp $t + 1$. The forget gate in the next cell controls if that information should be kept. It can either penalize or award the history by adjusting the weight based on the previous timestamp's cell states, hidden state, and current cell's input states. In the resident identification problem, if the residents' historical movement and activity patterns are not reviewed frequently, the forget gate tends to decrease the weight of those histories until they are negligible. If the pattern repeats frequently, the forget gate will fit the weight with a higher value. Thus, patterns that repeat frequently are preserved by the LSTM model, in contrast to Markov's assumptions that HMM models are based on that the current states only depend on the previous state. The LSTM's hidden state contains not only the state information about the previous timestamp but also all historical states. This enables the model to learn a longer sequence of historical activities by the LSTM model. It also has more model parameters that could be trained compared to statistical models; thus, more information could be learned. Because of this property, the LSTM model is selected to perform the resident identification task.
The bi-directional LSTM is chosen since it can consider more contextual information about the sequence of events. Both forward and backward contexts will be considered by the LSTM model. The bi-directional LSTM allows the model to consider the sequences in both directions. The added contextual information will provide a better performance than the unidirectional LSTM. The existence of backward relationships in the model could also allow it to make corrections when future events are added to the sequences. The future events included in the sequences give the model have better understanding and prediction of residents' spatial and temporal movement.


The data used to train and evaluate the LSTM model is the concatenation of the trained node embeddings and transformed time vectors in previous steps. Originally, each sensor event log record consisted of a sensor ID of the sensor triggered and the timestamp of this event recorded. After the data pre-processing, the timestamp is replaced by the time vector, and the node embedding vector of this sensor is concatenated to the end of the feature vector. Later, those features are fed into the LSTM model for training. The dataset used for this resident identification framework is a variable-length contiguous time series dataset. It is too long as a single input for the LSTM model. Meanwhile, LSTM models only handle fixed-length sequences as input. Therefore, the data needs to be split into fixed-sized chunks first. Intuitively, a chunk size that splits the long sequences into shorter ones approximates the average number of events per day may be a suitable choice. While the optimal number may differ from this strategy, it is evaluated and discussed in later sections.

\begin{table}
\centering
\caption{Example of TWOR2009 dataset}
\label{tab:twor2009_example}
\begin{tabular}{ c c c c c }
\toprule
time & sensor & status & event & begin/end \\
\midrule
08-24 00:00:19 & M050 & ON & R1\_Wandering & begin \\
08-24 00:00:19 & M044 & ON & & \\
08-24 00:00:21 & M046 & ON & & \\
08-24 00:00:21 & M044 & OFF & & \\
08-24 00:00:22 & M050 & OFF & R1\_Wandering & end \\
08-24 00:00:25 & M046 & OFF & R1\_Wandering & begin \\
08-24 00:00:41 & M044 & ON & & \\
08-24 00:00:43 & M044 & OFF & & \\
08-24 00:00:56 & M044 & ON & R1\_Wandering & end \\
\bottomrule
\end{tabular}
\end{table}

\section{Experiment \& Data}\label{sec:experiment}

We have used two real-world datasets to evaluate the proposed method. One publicly available dataset, known as the TWOR dataset, is collected from the Center for Advanced Studies in Adaptive Systems (CASAS) project \cite{twor_dataset}; the other dataset is collected from our smart office experiment setup. The datasets are separated into two partitions, 25\%  for testing purposes and 75\% for training purposes. It is essential to recognize the limitations of using a two-way split, as it may lead to overfitting the testing set during model selection. To mitigate this risk, we have employed techniques like cross-validation, allowing us to use the entire dataset for training and testing while still providing an unbiased performance estimate.  A randomized 10-fold cross-validation is applied. Models are compared with their average test performance of cross-validations.

\subsection{TWOR Dataset}

TWOR2009 dataset is collected at a sensor-rich smart home testbed known as Kyoto \footnote{Source of TWOR dataset: \href{http://casas.wsu.edu/datasets/twor.2009.zip}{http://casas.wsu.edu/datasets/twor.2009.zip}}. This testbed is a part of the CASAS project, and data were collected for a two-month period consisting of 138000 records. The testbed held two residents. It is a 2-level apartment with two bedrooms on the second floor, with a kitchenette, storage room, living room, and entrance on the first floor. Each room has multiple motion sensors installed; all doors are connected with door sensors, and all appliances can report their usage status. Each record of the dataset is a 4-tuple, including the timestamp of the event, name of the sensor, reading of the sensors respectably, and the annotation, indicating the identity and type of activities. Table \ref{tab:twor2009_example} is a snippet of the dataset that shows the attributes. The residents' activity events are labeled with their identity and activity.




\subsection{Smart Office Setup \& Data Collection}

\begin{figure}
    \centering
    \includegraphics[height=8.5cm]{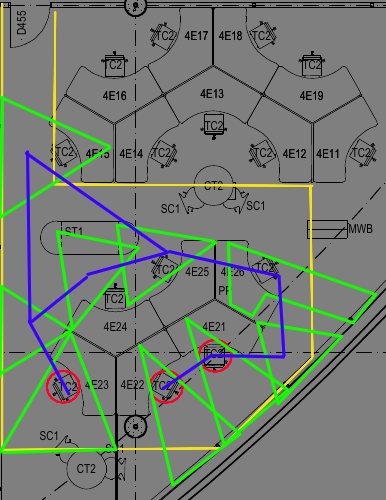}
    \caption{Experiment layout map for smart office setup.}
    \label{fig:exp_layout}
\end{figure}

The data collected from our smart office setup consists of 9 motion sensors and 3 test subjects. There is one sensor for each subject installed on their respective office cubicle. The rest of the sensors are installed in corridors connecting these cubicles and at the office entrance. The experiment was conducted for a contiguous two weeks only on weekdays, and a total of 10.5 days of sensor events logs were recorded. The data and annotation are arranged similarly to the TWOR dataset. Two thermal infrared cameras were utilized only for data annotations (i.e., creating ground truth). Annotation is done manually by observing the infrared camera recordings. The layout map of the office, the position of motion sensors, and assigned cubicles can be found in Fig. \ref{fig:exp_layout}. The three subjects' cubicles are shown as red circles; the yellow border indicates the boundary of the experiment; green shapes in the layout map illustrate the approximate detection area of each motion sensor; the blue network shows the generated AG. Two subjects ($R1, R2$) have adjacent cubicles assigned; the other subject ($R3$) is at the opposite cubicle of $R2$. There are other residents other than the three subjects in the office. Events caused by them are treated as noise and removed during the annotations process. In this experiment setup, we are using the center of the detection area of each motion sensor as POI for constructing the AG. Due to the confinement of the experiment, we cannot include a sensor to cover the corridor above 4E25 in Fig. \ref{fig:exp_layout}. The constructed AG by the proposed algorithm contains two subgraphs. To enable connectivity between the two sections of the office, an edge is manually added, crossing the barrier between the topmost cubicles on the right (4E25 and 4E26).

\begin{figure}
    \centering
    \includegraphics[width=10cm]{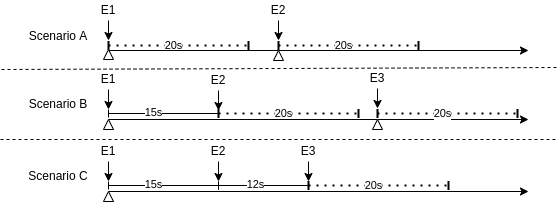}
    \caption{Timeline for illustrating event detection interval in off-the-shelf motion sensors.}
    \label{fig:detection_interval}
\end{figure}

The motion sensors we used are consumer-graded, off-the-shelf sensors. They were imposed with a \textit{detection interval}, believed to be an artificial limitation by sensor vendors for longer battery life and ease of automation setups for average home users. Those motion sensors cannot \textit{report} another event within a certain period of time. For instance, if a motion sensor has a 20-second detection interval. If such a sensor detects a movement for person A at $T+00:00$, it will report such an event once the movement is detected instantly. However, it will not report any event between $T+00:00$ to $T+00:20$ regardless. At $T+00:20$, the motion sensor will make another attempt to report its status. If movements happened within its range in the previous 20 seconds, the motion sensor would report a successful detection of movement at $T+00:20$. It only reports not detecting any movement after no movement has happened for the last 20 seconds. Fig. \ref{fig:detection_interval} demonstrates the movement timelines and how off-the-shelves motion sensors report activities. There are three scenarios showing the concept of detection interval. $E1, E2, E3$ are events made by residents within the detection range of motion sensors; the triangles below the timeline show when the motion sensor will report a detection; The dashed line shows the detection interval. The detection interval is considered one of the experiment's limitations. One common scenario is if one resident moves past a sensor, followed by another resident, within 20 seconds, only one detection would be reported. It is also possible that a person moves past a motion sensor and moves past the motion sensor again in the opposite direction during the detection interval. In this case, only one event is reported. Thus, resident identification models may expect the resident at the new location, although the resident has returned to the initial location. We believe our method would perform better without this limitation. 



A down-sampling and up-sampling process is applied to the dataset before training the model. Testing subjects often stayed stationary, usually being in their own cubicle for a long period each day. In this experiment, stationary events contribute 90\% to the recorded data. In this context, the trained model is biased toward stationary events. This imbalance of data made models tend to predict all sensor events belonging to one subject being stationary at their cubicle. The down-sampling process introduced a rate limit for continuous stationary sequences of events for subjects staying at their \textit{home sensor}. A \textit{home sensor} is defined as the sensor installed in the subjects' respective cubicles. The down-sampling process attempts to limit the number of continuous sequences of home sensor events. It is defined as, within a time interval $T$, if a home sensor detects more than one continuous event, only the first event is kept, and all other events made by such home sensors are removed. Occasionally, subjects will wander into the office or go out of the office. The down-sampling process reduces the weights on stationary data and adds more weight to movement data. where positional encoding proposed in this paper could benefit the most. After the down-sampling process, there are 200 sensor events on average per day. An up-sampling process is applied to the training dataset for a larger amount of data used for training. The up-sampling duplicates the data after chunking and train-test split of the data. For this particular dataset we collected, the training dataset is duplicated by 8 to increase the information to be trained per epoch by the LSTM model. With more data included in the training dataset, we can observe increases in performance with fewer epochs.

\subsection{Research Questions}

For the two datasets, multiple training parameters are selected for evaluation. This includes the volume of sensor events, with or without down/up-sampling, and multiple hyper-parameters related to the Node2Vec embeddings, which may impact the performance of node embeddings. A baseline model has been set up for reference. For both datasets, two baseline models are included, the one without any positional encoding information included and the one with coordinates as positional encoding. For the TWOR dataset, an extra baseline model, which uses room numbers of sensors as positional encoding, is included. For the dataset we collected, the experiment was split into four blocks. The first two blocks use only half of the data available, i.e., five days in our experiment. The other two blocks use the entire available dataset. Within each of the two blocks, one is conducted with down/up-sampling of the events, and the other is without. The objective is to test whether the up/down-sampling process benefits the model and how much it does. The hyper-parameters of the Node2Vec algorithm are tested on the TWOR dataset. The effect of how the volume of data affects the models' performance is evaluated by comparing the model trained by the full dataset and by subsets of data on the TWOR data. Models' performance is also prepared between the TWOR dataset and the dataset collected by our experiment. To address the above problems, we design, evaluate, and discuss experiments and come out with the following research questions:

\begin{itemize}
    \item \textbf{RQ1}: How do the sequence length and the number of random walks of the Node2Vec algorithm influence the model's performance?
    
    \item \textbf{RQ2}: What is the impact of the dimension size of the Node2Vec algorithm with the model? Does a larger dimension size lead to extra computation time?

    \item \textbf{RQ3}: What is the effect of window size of Node2Vec algorithm in the resident identification problem? what is the optimal value of this parameter?

    \item \textbf{RQ4}: How does the chunk size of the LSTM model affect the overall performance? What is the optimal chunk size?

    \item \textbf{RQ5}: What is the impact on the volume of data used for training? Does a smaller data volume still produce acceptable performance?
\end{itemize}

\section{Results \& Discussions}\label{sec:result_discussion}

Four parameters of the Node2Vec algorithm may affect the node embeddings' performance on the resident identification approach. They are the dimension size, the sequence length, the number of random walks, and the window size of the Node2Vec algorithm. The dimension size is the vector length of the output node embeddings. Generally, the larger the dimension size is, the more information is included by the embeddings. However, a large dimension size might have a negative impact on the performance of machine learning models. The number of random walks and the length of random walks control the volumes of contextual and topological information to be included. The sample size of the underlying Word2Vec model is controlled by the number of random walks of the Node2Vec algorithm. The random walk of the graph is a process of sampling the graph to generate the required sequence that is required by Node2Vec's underlying Word2Vec algorithm. If there is insufficient sequence length and the number of random walks, it is possible that some nodes are not traversed, and some graph sequence patterns are not learned. The resulting node embeddings are likely to be low-performing. However, using large numbers on these two parameters might be too computationally difficult. The window size of the Word2Vec algorithm controls how the model tries to associate neighbors of words with itself. In the Node2Vec algorithm, it is the parameter controlling how many neighbors should be considered as contextual to positions. The effect of this parameter is discussed in a later sub-section.

During the evaluation process, accuracy, precision, recall, and F1 score are selected for comparing the models' performance. Mainly, the accuracy score is used. The epoch selects the best model with the lowest validation loss. All experiment setups are run for at least ten repeats to eliminate the probability of the randomness effect. Statistical tests are employed when necessary. \emph{The source codes are available in the GitHub repository to evaluate the correctness of the approach and for easy reproducibility.} \footnote{GitHub repository for the experiment: \href{https://github.com/Song-Zhiyi/Resident-Identification-with-Positional-Encoding}{https://github.com/Song-Zhiyi/Resident-Identification-with-Positional-Encoding}}.



%


\subsection{Effectiveness of the Proposed Model}

\begin{figure*}[htbp]
    \centering
    \subfloat[TWOR dataset.]{
        \includegraphics[width=7cm]{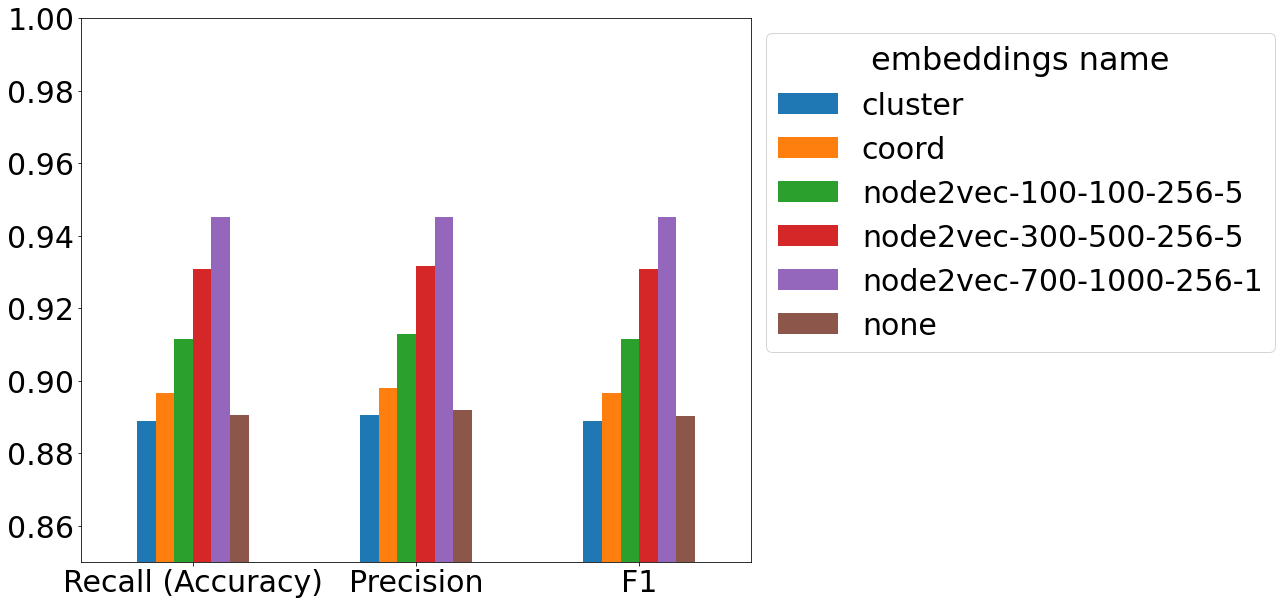}
    }
    \subfloat[Our dataset.]{
        \includegraphics[width=7cm]{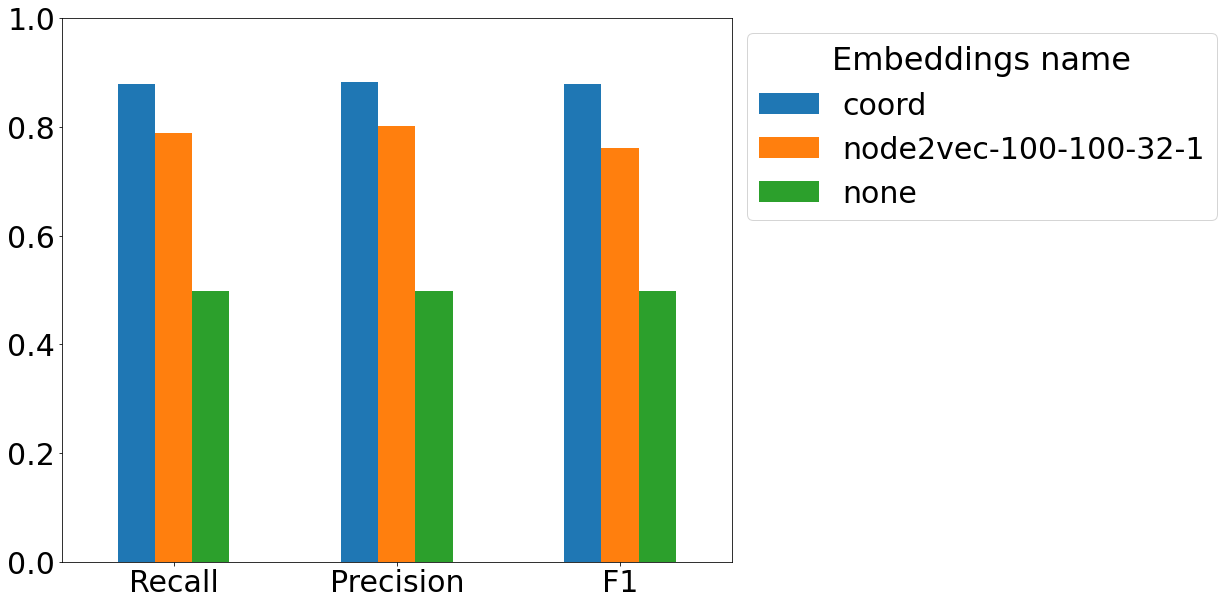}
    }
    \caption{Comparison of performance metrics.}
    \label{fig:result}
\end{figure*}

We observed overall increases in accuracy scores compared to baseline models. Fig. \ref{fig:result} shows the comparison of the model performance of all experiments. For the TWOR dataset, the accuracy score of the first baseline model without positional encoding is 89.0\%. By using coordinates as positional encoding, the second baseline model provides 89.7\% accuracy. The second baseline model shows a statistically significant improvement over the first baseline model. However, when using the room number of sensors as positional encoding in the third baseline model, the model is not observed with performance improvement over the model without positional information included, nor is it statistically significant. This method does not provide improvement as the feature space may be too small to include enough positional information. The proposed method, with the Node2Vec node embeddings included, the best model using our proposed method could reach 94.5\% accuracy on the TWOR dataset. It is statistically significant. Node embeddings with other complexity have performance varying based on the characteristics of 4 hyper-parameters of the Node2Vec algorithm. The detailed effect of those parameters is discussed in a later section.  

For our dataset, the model performed poorly without any positional information; the baseline model is 45\% for data without down/up-sampling and 50\% for sampled data. We only observed significant model performance improvement on the sampled dataset for the proposed method. This suggests that the down-sampling process effectively reduces the imbalance between stationary and movement data. The best node embeddings with a random walk size of 200, a sequence length of 200, a dimension size of 128, and a window size of 1 have an accuracy score of 88.4\%. However, surprisingly, the model with coordinates as positional information performs exceptionally well on this dataset, reaching an accuracy of 87.9\%. This is not within our initial expectations. The model using Node2Vec embeddings only shows slight improvement over the coordinates-based positional encoding. It may be the confinements of the experiment, where the experiment is conducted in a small confined area with off-the-shelf devices. The home structure is simple; thus, a simple positional encoding, like coordinates, could improve the performance of the model. Another reason for this phenomenon is the extra dimensions of the positional encoding of the proposed method compared to the simple one. Considering the topology of our setup is simple, the extra dimensions produced by the proposed method provide a limited improvement to the resident identification model. Coordinates may work better in this scenario because the lower dimension is because of the simple topology and lesser dimensional included. However, we believe this phoneme only applies to very simple typologies. With complex topologies, the proposed method of generating positional encoding will help the resident identification model better understand of the residents' movements.


\subsection{Complexity of Node Embedding}

The performance impact of hyper-parameters is tested and evaluated by controlled experiments. The model with 700 random walks of the APG, 1000 sequence lengths of the random, 256 as dimension size, and 5 as window size of the Node2Vec model are selected as the baseline model for evaluation of node embeddings' hyper-parameters.


\begin{figure*}
    \subfloat[Number of random walk, sequence length.]{
        \includegraphics[width=4.6cm]{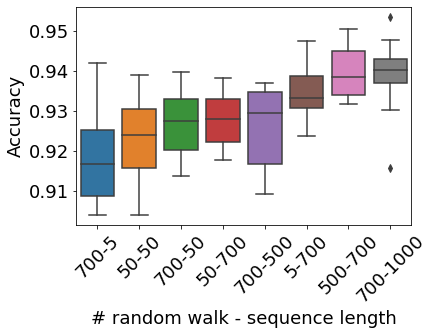}
        \label{fig:result_seq_len_walk_n_together}
    }
    \subfloat[Dimension size.]{
        \includegraphics[width=4.6cm]{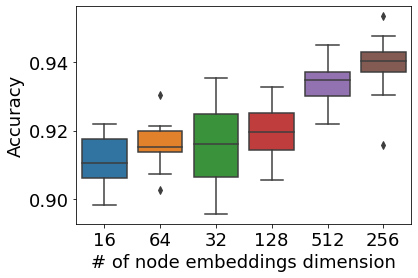}
        \label{fig:result_dimension_size}
    }
    \subfloat[Window size.]{
        \includegraphics[width=4.6cm]{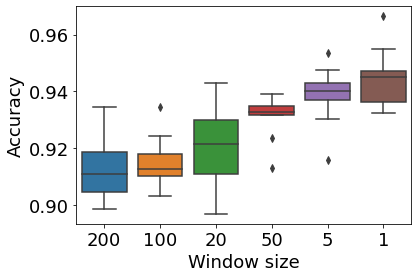}
        \label{fig:result_window_size}
    }
    
    \subfloat[Chunk size.]{
        \includegraphics[width=4.6cm]{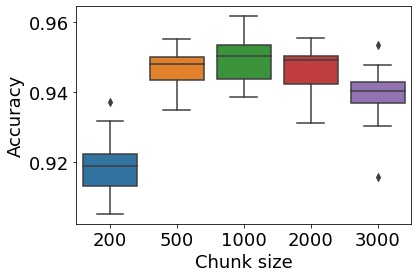}
        \label{fig:result_chunk_box}
    }
    \subfloat[Volume of data.]{
        \includegraphics[width=4.6cm]{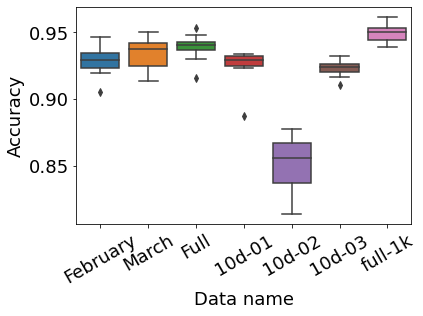}
        \label{fig:result_duration}
    }
    \subfloat[Comparison of accuracy on volume of data for both dataset.]{
        \includegraphics[width=4.6cm]{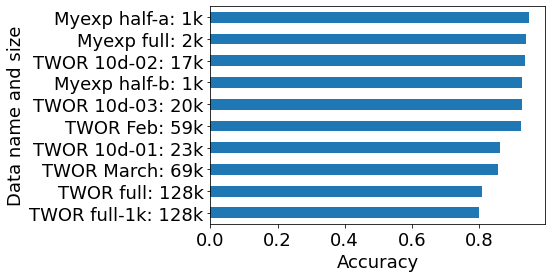}
        \label{fig:result_myexp_duration}
    }
    \caption{Comparison of performance on parameters based on accuracy.}
    \label{fig:exp_result_1}
\end{figure*}

\subsubsection{Sequence Length and Number of Random Walks (RQ1)}

Fig. \ref{fig:result_seq_len_walk_n_together} shows the performance comparison of the two parameters. Both variables seem to control the volume of information about the original graph included in node embeddings. However, a lower limit on sequence length may exist to produce node embeddings with acceptable performance. By observing two node embeddings training setup, with the number of random walks and sequence lengths being 700, 5, and 5, 700 respectively, the embeddings with larger sequence lengths outperformed significantly their counterparts. For another two embeddings with two parameters being 700, 50, and 50, 700, respectively, no significant difference could be observed. By contrasting the two pairs, we conclude a minimal sequence length, at least at the scale of the number of sensors in the AG, is required. Sequence length less than this minimal length may include not enough information about the connectivities of the AG, thus producing poorly performed node embeddings. The influence of these two parameters on models' performance is within expectation. They have a positive correlation with performance and training time. However, it is unlikely for the sensors' location to change after installation, and retraining of the embeddings may not be frequent. The training time for node embeddings is far less than for fitting and evaluation of resident identification models. In this regard, a higher value on these two parameters is suggested to maximize the information extracted from the AG.


\subsubsection{Dimension Size (RQ2)}

The dimension size of Node2Vec controls the size of output vectors representing POI in an AG. It controls the volume of information to be included in the generated node embeddings. However, a large dimension size will add too much noise to the resident identification models. For the TWOR dataset, a dimension size of 256 seems to be the best selection, as shown in Fig. \ref{fig:result_dimension_size}. As for dimension sizes, it is statistically better than 16, 32, 64, and 128. For this dataset, which contains 62 sensors, a dimension size of 256 is about 4 times the number of the nodes in the AG, which has the best performance overall for all other selections on dimension size. Dimension size lower than 128 seems to be too small for the node embeddings to hold enough topological information for the resident identification task. The dimension of 512 seems to have a negative effect on the resident identification model. This may be due to the high dimension size of the input vector to the model.

\subsubsection{Window Size (RQ3)}

From Fig. \ref{fig:result_window_size}, we conclude a negative correlation between window size and the performance of node embeddings. Node embeddings with smaller window sizes have better performance. The window size parameter controls how many contextual nodes should be considered relevant to the center node. In Word2Vec embeddings, a larger window size makes the model believe more neighboring words have similarities; in Node2Vec models, more neighboring nodes are considered relative to the center node. This is not ideal for identifying movement patterns. Since the vertices of the AG are connected by possible pathways between POIs, relating non-adjacent POI to center POI would add noise to the node embeddings. Thus, we are observing a decreased accuracy score on resident identification. For resident identification tasks, the lower window size leads to better performance. However, considering occasional malfunctioning is common in consumer-graded motion sensors, the residents may pass a few sensors without triggering them. A slightly larger window size could add robustness to counter this issue. For example, using 5 as window size, the Node2Vec embeddings will associate at most 5 sensors away from the central sensor. The system could tolerate at most 5 continuous sensor failures in this case. The window size selection should depend on the home's complexity and quality of sensors. In ideal setups, a window size of 1 is optimal.

\subsection{Chunk Size of LSTM Model (RQ4)}

\begin{table}[!t]
    \centering
    \caption{Pair-wise comparison for chunk size}\label{tab:result_chunk_tukeyhsd}
    \begin{center}
        \begin{tabular}{ccccc}
            \toprule
            \textbf{group1} & \textbf{group2} & \textbf{meandiff} & \textbf{p-adj} & \textbf{reject} \\
            \midrule
            500             & 1000            & 0.0027            & 0.8439         & No              \\
            500             & 2000            & -0.0001           & 1.0            & No              \\
            500             & 3000            & -0.007            & 0.0741         & No              \\
            1000            & 2000            & -0.0027           & 0.8334         & No              \\
            1000            & 3000            & -0.0096           & 0.0065         & Yes             \\
            2000            & 3000            & -0.0069           & 0.0784         & No              \\
            \bottomrule
        \end{tabular}
    \end{center}
\end{table}

From Fig. \ref{fig:result_chunk_box}, we can observe a chunk size of 200 has the lowest performance. The tendency of model performance is first increased by chunk size and then decreases, which is within the initial assumption. Statistical testing is performed to distinguish the performance since the accuracy of the two groups is close. Table \ref{tab:result_chunk_tukeyhsd} is a TukeyHSD pair-wise test table. It shows the mean difference and p-value of hypothesis testing between groups. The statistic test suggests that a chunk size of 1000 performs best over other selections with an accuracy score of 95\%. A chunk size of 1000 is approximately $\frac{1}{3}$ of average sensor events per day. However, we cannot find any statistical difference between chunk sizes of 500, 1000, and 2000. This may indicate that the fine-tuning of chunk size for datasets depends on the dataset density of sensor events, and the number of overall sensor events may be required to gain the best model performance of the proposed method. In general, chunk size at the scale of a day is recommended.

\subsection{Volume of the Dataset (RQ5)}

This research question is designed to study how the volume of data affects the performance of the proposed framework. The proposed method is evaluated with a limited set of data. Since the current approach still relies on the annotation of data, we want to find the minimum amount of annotated data that would produce satisfactory results. Both datasets were evaluated to better understand how the volume of data affects the proposed method. For the TWOR dataset, two setups are tested. One setup splits the original data into two halves, each including one month of data (February, March); another setup splits the data into three 10-day chunks (10d-01, 10d-02, 10d-03). The resident identification models are trained with the same node embeddings. The result is shown in Fig. \ref{fig:result_duration}. The February and March data are not statistically different from the full data. Among the three 10-day chunked data, 10d-01 and 10d-03 have similar performance, whereas 10d-02 has an abnormally low performance. It may suggest significant differences in daily patterns during the 10-day period. For our dataset, there are two setups of experiments. The result is shown in Fig. \ref{fig:result_myexp_duration}. The first set of experiments uses the full dataset (i.e., Myexp full), and the other uses half of the dataset (i.e., Myexp half-a, Myexp half-b). It suggests the same result as the TWOR2009 dataset, where the experiment with fewer data does not show significantly worse results over full data. This is within the expectation. Residents usually have weekly patterns of activities. Therefore, if a dataset could include at least a week span of data, it should have acceptable performance if the daily pattern between weeks does not have significant differences. However, this may not be true for weeks apart (i.e., a week in summer and a week in winter). Overall, with 10 days of annotated data in both experiments, the performances are not significantly different from the full-length models. However, it still requires manual labor to annotate the data, and it may not be feasible in all real-world scenarios. It is one of the limitations of our proposed method, and so are all supervised learning methods. Nevertheless, the proposed framework still has its potential for the resident's identification problems. Since the APGs do not depend on the annotations and are not difficult to construct, they can be utilized in unsupervised models to increase the topological connection between continuous events.


Experiments on both datasets yield acceptable performances. Significant performance improvement is observable with the proposed resident identification framework. The best model for the TWOR dataset achieves 94.5\% accuracy in predicting residents' identity in a 2-resident home. And for our dataset, the best model has 88.4\% accuracy in a 3-resident office. It is also found that a larger sequence size, random walk, and dimension size of the Node2Vec algorithm leads to better performance. But for dimension size, being too large may introduce noise to the model. Thus, a reasonable proportion to the number of POIs is recommended to increase performance and save computational efforts. A window size of 1 is optimal for the resident identification task. However, a larger number adds robustness to the proposed framework, considering the possible malfunctioning of sensors. Multiple reasons could contribute to the gap in performance between the TWOR dataset and our experiment. The TWOR dataset has a denser sensor setup than our experiment. This could reduce blind areas that motion sensors could not cover; each sensor could be responsible for a smaller region, which adds more details to the AG. Their experiments have 50 motion sensors installed compared to 9 motion sensors in our experiment. Since their dataset is believed to be collected by motion sensors not having detection intervals, they are not suffering from problems introduced by detection intervals. A third factor might be the TWOR dataset is only for two residents, whereas ours is for three subjects.


\section{Conclusion}\label{sec:conclusion}

Positional encoding is crucial in smart home research where location awareness is required. The model's performance on the resident identification task improved because of the additional encoded topological information and structural information of smart environments. We proposed a novel way of encoding the positions of interest of homes into machine learning-friendly node embeddings. The method uses available information from the smart environment setup, typically a layout map. An algorithm is designed to build an Accessibility Graph and later transform the graph into an Accessibility Probability Graph (APG). The generated APG is used for the Node2Vec algorithm to produce node embeddings that encode POIs into vector form. The LSTM model was used for resident identification on the process data. The proposed methods are evaluated against two datasets: a dataset publicly available and a dataset collected from our own testbed. Significant performance improvement is observed in proposed positional encoding over baseline models.

To solve the problem that all supervised learning has, the need for labeled data, we are trying to explore techniques like transfer learning, in which we can train generalized models and fine-tune them in individual smart environments. The pre-training of models has become a trend in many areas. The same concept could be used in smart home problems. It is also possible to include some reinforcement learning into the model so that that model could adjust itself based on positive and negative feedback from residents.
In the future, the Seq2Seq model could be adjusted from an N-to-N Seq2Seq model into an N-to-1 model. So that real-time prediction of residents' identity would be possible. To predict which residents are responsible for an event captured by an environmental sensor, we can feed sequences of sensor events till the latest event into the LSTM model. The N-to-1 LSTM model will predict the identity of the residents of the latest event by considering all historical temporal and spatial traces of residents provided by the sensor event sequences. This model will take a sequence of N events where the last event is the latest event captured in real-time as input and produce a single output to indicate the resident's identity. The current approach does not consider guests. Since the movement of guests captured by the environmental sensors may be similar to the residents, it may be considered as one of the residents' movement traces. There is no good solution for a current supervised approach. This problem should be addressed in future models, where the habit patterns of residents could be extracted. For those models, the proposed positional encoding would be useful to boost their effectiveness because of the included positional information.

The proposed method that builds the node embeddings representing the home's structure may be used in areas other than resident identification. In the future, it is possible to generalize to other problems that require topological awareness of locations and connectivities of locations in homes and the structure of homes (i.e. habit pattern extraction, etc.). For those questions, similar approaches to constructing node embeddings could be applied. During the experiment, we only encoded the location of sensors using this method. While the method is not limited to the location of sensors, all POIs could be encoded in such a way. For other problems that require topological or structural information on locations in homes, the proposed positional encoding method might be beneficial as well. Because of the added information about possible pathways of homes and how location is connected, the models' performance might be increased.

\bibliographystyle{ACM-Reference-Format}
\bibliography{References}


\end{document}